\documentclass[runningheads]{llncs}
\usepackage[T1]{fontenc}
\usepackage{caption}
\usepackage{subcaption}
\usepackage{amsmath}
\usepackage{booktabs}
\usepackage{pifont} 
\usepackage{multirow}
\usepackage{ amssymb }
\usepackage{siunitx}
\usepackage{pifont}
\newcommand{\cmark}{\ding{51}}%
\newcommand{\xmark}{\ding{55}}%
\usepackage{hyperref}
\hypersetup{
    colorlinks=true,
    linkcolor=blue,
    citecolor=blue,
}
\usepackage{graphicx}
\usepackage{soul}
\usepackage{multicol}
\usepackage{multirow}
\usepackage{graphicx,verbatim}
\begin{document}
\title{Anatomically-guided masked autoencoder pre-training for aneurysm detection}
%

\author{Alberto M. Ceballos-Arroyo\inst{1} \and Jisoo Kim  
\inst{2,4} \and Chu-Hsuan Lin \inst{2} \and Lei Qin \inst{3,4} \and \\ Geoffrey S. Young \inst{2,4} \and Huaizu Jiang \inst{1} 
}

\authorrunning{Ceballos-Arroyo et al.}

\institute{Northeastern University\\
\email{\{\textbf{\underline{ceballosarroyo.a}},h.jiang\}@northeastern.edu}\\ \and
Brigham and Women's Hospital 
\{\email{jkim,clin71,gsyoung\}@bwh.harvard.edu}\\
\and
Dana-Farber Cancer Institute
\{\email{lei\_qin@dfci.harvard.edu}\}\\
\and
Harvard Medical School
}

\maketitle             
\begin{abstract}

Intracranial aneurysms are a major cause of morbidity and mortality worldwide, and detecting them manually is a complex, time-consuming task. Albeit automated solutions are desirable, the limited availability of training data makes it difficult to develop such solutions using typical supervised learning frameworks. In this work, we propose a novel pre-training strategy using more widely available unannotated head CT scan data to pre-train a 3D Vision Transformer model prior to fine-tuning for the aneurysm detection task. Specifically, we modify masked auto-encoder (MAE) pre-training in the following ways: we use a factorized self-attention mechanism to make 3D attention computationally viable, we restrict the masked patches to areas near arteries to focus on areas where aneurysms are likely to occur, and we reconstruct not only CT scan intensity values but also artery distance maps, which describe the distance between each voxel and the closest artery, thereby enhancing the backbone's learned representations. Compared with SOTA aneurysm detection models, our approach gains +4-8\% absolute Sensitivity at a false positive rate of 0.5. Code and weights will be released.


\end{abstract}

\section{Introduction}

Intracranial aneurysms (IAs) are outpouchings that occur in brain arteries and occur due to the weakening of the vessel's wall. Aneurysms can rupture under certain conditions, leading to potentially deadly sub-arachnoid hemorrhage (SAH), a condition with a 30-day mortality rate of 40-50\%~\cite{Park_Lee_Heo_Han_Lee_Hong_Lee_Lee_Oh_2022}. Expert radiologists are capable of detecting most aneurysms through visual inspection of patient scans, but such is a time-consuming process involving the manual inspection of hundreds of images. As a result, the research community has striven to develop automated solutions that can assist clinicians in detecting aneurysms. Most such solutions are deep-learning-based, with some achieving over 90\% Sensitivity with False Positive (FP) rates below 2 per scan \cite{ceballos-arroyoVesselAwareAneurysmDetection2024,wangDetectionIntracranialAneurysms2023}. 

Despite ongoing efforts, many of these aneurysm detection models perform significantly worse on out-of-distribution testing data (\textit{i.e.}, scans taken at different medical centers and/or with different equipment than the training data) \cite{boHumanInterventionfreeClinical2021,ceballos-arroyoVesselAwareAneurysmDetection2024,shiClinicallyApplicableDeeplearning2020}, restricting their clinical applicability. The limited generalization capability of some aneurysm detection approaches can be linked to the strictly supervised nature of their training pipelines \cite{boHumanInterventionfreeClinical2021,shiClinicallyApplicableDeeplearning2020,dinotoAutomatedBrainAneurysm2023}. While supervised learning can be enough for some tasks, it requires annotated data, which is not readily available for aneurysm detection. For instance, the largest annotated public IC dataset \cite{boHumanInterventionfreeClinical2021} has a training partition of 1,186 CT scans, 
and most recent work \cite{assisAneurysmPoseEstimation2023,dinotoAutomatedBrainAneurysm2023} relies on even smaller datasets (100-500 scans) for training. Thus, such models are prone to over-fitting to the data distributions present therein. 

\begin{figure}[t]
\centering
\includegraphics[width=1\textwidth]{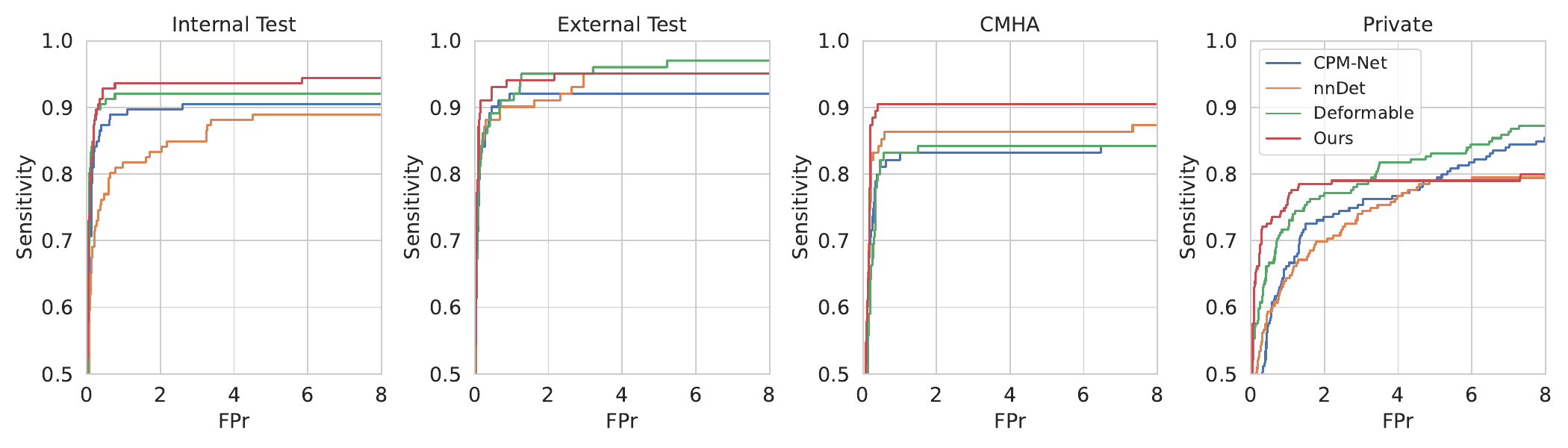}
\caption{Lesion-level Sensitivity vs FP rate curve for our best model compared with three baselines, measured across four datasets. Our model (red) consistently achieves better Sensitivity for tolerances of 0 to 2 FPs per scan, which are essential to minimize the amount of time radiologists spend reviewing FPs. }
\label{fig:froc} 
\end{figure}

Pre-training is a popular choice when data for a task is limited. This approach can help models generalize in the natural images domain \cite{dengGeneralizationAbilityUnsupervised2024,heMaskedAutoencodersAre2022}, it is not frequently used for aneurysm detection, however, as most pipelines rely on CNN-based backbones \cite{isenseeNnUNetSelfconfiguringMethod2021,baumgartnerNnDetectionSelfconfiguringMethod2021a}.
While there are various pre-training methods for CNNs, some require access to annotated datasets like ImageNet 
\cite{ridnikImageNet21KPretrainingMasses2021}, with no direct equivalent in the medical domain; others, like augmentation-based contrastive learning \cite{caronEmergingPropertiesSelfSupervised2021b}, rely on applying color or spatial transforms that may not result in realistic data for modalities like CT. 
An approach which better fits the characteristics of CT data while still enabling better generalization is masked-autoencoder pre-training (MAE) \cite{heMaskedAutoencodersAre2022}. 
In specific, a Vision Transformer (ViT) \cite{dosovitskiyImageWorth16x162021a} model is used to mask out portions of unlabeled images which are then reconstructed by a decoder. This method has been successfully leveraged to pre-train ViTs across various domains \cite{congSatMAEPretrainingTransformers2023,zhouSelfPretrainingMasked2023} without relying on augmentations or annotations. Moreover, ViTs have shown to perform even better than CNNs in some instances \cite{dosovitskiyImageWorth16x162021a}. However, there are several particularities to consider: first, the amount of data required for pre-training is often larger than for supervised training; second, MAE was designed for pre-training on 2D datasets using large batch sizes, but the volumetric nature of aneurysms and CT scans necessitates 3D attention, which might limit throughput if used naively; finally, CT data has a degree of sparsity, with parts of the scans being irrelevant to aneurysm detection, limiting the usability of existing 3D MAE approaches \cite{tongVideoMAEMaskedAutoencoders2022,feichtenhoferMaskedAutoencodersSpatiotemporal2022}.

To address the issues above, we propose a self-supervised pre-training pipeline consisting of several key components. 
First, we collect a dataset comprising 6,796 head CT scans from various source, from each of which we are able to crop dozens of samples for pre-training.
Second, we enable reasonably fast pre-training of a 3D ViT by using a factorized 3D self-attention mechanism inspired by the video understanding literature \cite{arnabViViTVideoVision2021a}.
Third, we modify MAE pre-training to better fit aneurysm detection by using a sampling mechanism for crops and masked patches that emphasizes anatomically relevant regions, \textit{i.e.}, those that are close to brain arteries. Moreover, instead of solely reconstructing CT scan intensity values, we use continuous artery distance maps, which describe the distance of each voxel to the nearest artery, as a second input channel for our pipeline, forcing the model to associate certain visual patterns with proximity to blood arteries. For both kinds of inputs, we minimize the Mean Squared Error (MSE). The pre-trained model is then used as the feature extractor for our detection pipeline, where the decoder is query-based Transformer inspired by DETR \cite{zhuDeformableDETRDeformable2021}. We highlight the following contributions of our work:

\begin{itemize}
    \item We propose an entirely Transformer-based pipeline for 3D aneurysm detection relying on self-supervised pre-training on a collection of 6,796 CT scans, encompassing low- and high-resolution scans from various institutions.

    \item We improve on basic masked auto-encoder training by introducing anatomical information into the pre-training step both in the features to be reconstructed and the way sampling is carried out.

    \item Our pipeline matches state-of-the-art Sensitivity when evaluated on in- distribution data and surpasses it on out-of-distribution data by a margin of 4-8\% absolute Sensitivity given a tolerance of at most 0.5 FPs per scan.

\end{itemize}

\section{Materials and methods}

\subsection{Data}

For pre-training, we use head CT scans from various public datasets \cite{chilamkurthyDeepLearningAlgorithms2018,hooperImpactUpstreamMedical2021a,khoruzhayaExpandedBrainCT2024,yangBenchmarkingCoWTopCoW2024}, including low- and high-resolution samples. Table \ref{tab:data} describes our  data sources in detail. In total, we use 6,796 scans for pre-training, of which 1,391 are medium to high resolution (slice thickness between 0.3 and 1 mm) and the rest are low resolution (slice thickness greater than 1 mm). As none of the pre-training datasets include data annotated for aneurysm detection, we use both training and testing data where available. We finetune our model on Bo \textit{et al.}'s \cite{boHumanInterventionfreeClinical2021} dataset, whose training partition contains over 1,000 scans from various clinical centers in China. Bo's dataset also contains two evaluation partitions, one internal from the same clinical centers as the training data, and another external from a different set of hospitals. In addition, we use the CMHA dataset  \cite{songIntracranialAneurysmCTA2024} and our own privately annotated partition using data sourced from the United States, following hospital guidelines for acquisition and use. Dataset statistics are provided in Table \ref{tab:data}.

\begin{table}[h!]
\fontsize{8pt}{9pt}\selectfont
\centering
\caption{ Summary of the head CT datasets used in this paper.  }\label{tab:data}
\begin{tabular}{|l|l|l|l|c|c|}
\hline
\textbf{Dataset} & \textbf{Origin} & \textbf{Purpose} & \textbf{Resolution} & \textbf{\# scans}& \textbf{\# aneurysms }\\
\hline
Sino-CT \cite{hooperImpactUpstreamMedical2021a} & China & Pre-training & Low & 5,406 & -- \\ 
CQ500 \cite{chilamkurthyDeepLearningAlgorithms2018} & India & Pre-training & Medium-High & 396 & -- \\
MosMED \cite{khoruzhayaExpandedBrainCT2024} & Russia & Pre-training & High & 870 & -- \\
TopCoW \cite{yangBenchmarkingCoWTopCoW2024} & Switzerland & Pre-training & High & 125 & -- \\ 
\hline
Bo, Int. Train \cite{boHumanInterventionfreeClinical2021} & China & Fine-tuning & High & 1,186 & 1,373 \\ 
\hline 
Bo, Int. Test \cite{boHumanInterventionfreeClinical2021} & China & Evaluation & High & 152 & 126 \\ 
Bo, Ext. Test \cite{boHumanInterventionfreeClinical2021} & China & Evaluation & High & 138 & 101 \\
CMHA \cite{songIntracranialAneurysmCTA2024} & China & Evaluation & High & 141 & 95 \\
Private & United States & Evaluation & High & 143 & 219 \\
\hline 
\end{tabular}
\end{table}

For both pre-training and fine-tuning, we resample all scans to have a spacing of 0.4 mm and we crop areas below the skull base. We follow \cite{ceballos-arroyoVesselAwareAneurysmDetection2024,yadavDynamicComputedTomographyAngiography2025a} to train a nn-UNet \cite{isenseeNnUNetSelfconfiguringMethod2021} model specifically for artery segmentation rather than vessel segmentation (we exclude veins due to venous aneurysms being rare and less clinically relevant \cite{hoellCorticalVenousAneurysm2004}). Our artery segmentation model was trained on a set of 25 dynamic CT scans annotated with iCafe~\cite{chenDevelopmentQuantitativeIntracranial2018}, and achieved a modified Dice Score ([$Y$ $\cap$ $\hat{Y}$]/$\hat{Y}$) of 0.89 on a held-out validation set. Using the nn-UNet, we generate artery distance maps using the signed distance transform as described in \cite{ceballos-arroyoVesselAwareAneurysmDetection2024}. Since the labels for all datasets are dense 3D segmentations, we convert them into tight bounding cubes to accommodate our detection pipeline.

\subsection{Modeling}

\subsubsection{Transformer Encoder.}

Our encoder follows the 3D ViT architecture \cite{dosovitskiyImageWorth16x162021a}, with a sinusoidal 3D positional embedding. 
The input patch size is $2\times 64\times 64\times 64$. The first channel consists of intensity values from the CT scan and the second channel is an artery distance map that encodes how far each voxel is from the nearest artery. To ensure 3D attention fits within our computational budget, we factorize self attention by computing it in two steps.
For every Transformer layer, we first compute self-attention among tokens within the same slice, and then we compute the attention across slices by having tokens attend only to their temporal neighbors. In addition, a learnable \textit{CLS} token is able to attend to all tokens during both steps to better encode global information.  Thus, the total memory complexity of computing attention is bound by $O(N^2)$, where $N$ is the number of tokens in each slice. 

\subsubsection{Detection module.}

Given a set of tokens (3D features) from the encoder, we use a 3D sinusoidal positional encoding to introduce positional information. Next, we define $n_q$ learnable queries with the same feature dimension as the encoder's output as in~\cite{detr}. Then, we use cross-attention with the 3D features as key and values followed by self-attention among the queries. After this step, we feed the attended queries through three MLP heads, where each query is responsible for predicting the class (whether it is an aneurysm or not), location, and size for an aneurysm (can be empty meaning no aneurysm detected for a query). Since the number of queries is relatively small, we are able to use full attention in the detection module.

\begin{figure*}[t!]
    \centering
    \begin{subfigure}[t]{0.15\textwidth}
        \centering
        \includegraphics[height=0.8in]{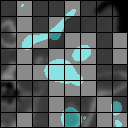}
        \caption{}
    \end{subfigure}%
    ~ 
    \begin{subfigure}[t]{0.85\textwidth}
        \centering
        \includegraphics[height=1.75in]{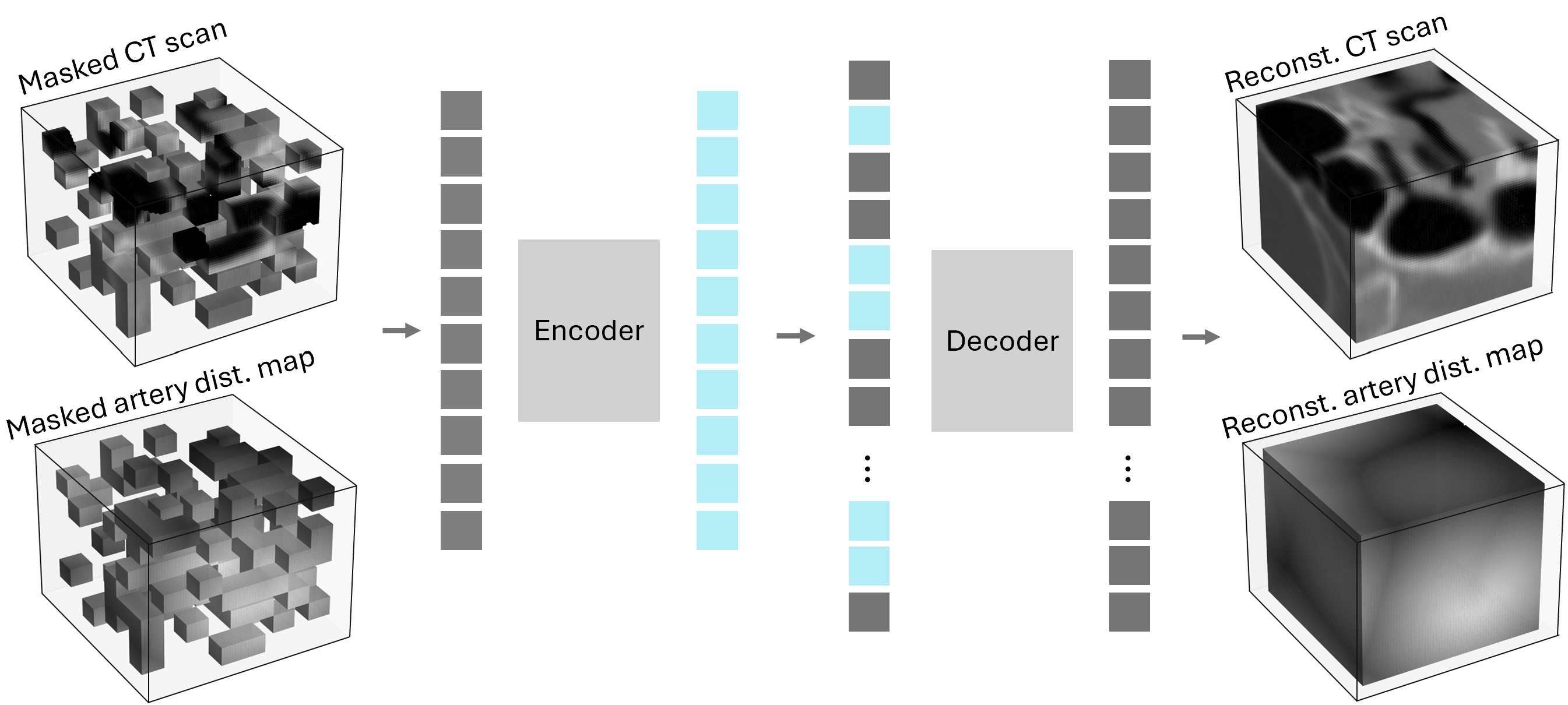}
        \caption{}
    \end{subfigure}
    \caption{(a) Visual depiction of our masking scheme on a single CT scan slice: lighter patches, which mostly overlap with vessels (cyan) areas, are masked; the model can only see the darker areas during pre-training. (b) Illustration of our MAE pipeline, with both the CT scan and the distance map being reconstructed. }
    \label{fig:pipe} 
\end{figure*}

\subsubsection{Pre-training strategy.}

To enhance representation learning and thus improve the generalization ability of our model, we do pre-training for the encoder. 
To this end, we add an auxiliary decoder, as shown in Fig.~\ref{fig:pipe} (b). 
Our pre-training method builds on the MAE strategy, where patches in a given sample are randomly masked out and a decoder is forced to reconstruct them based on the intermediate representation produced by the encoder and the architecture is optimized by minimizing the MSE for the masked tokens. After pre-training, the decoder is discarded and the encoder can be fine-tuned on related downstream tasks \cite{heMaskedAutoencodersAre2022}. 

In order to match the input size defined earlier, we randomly crop $64\times 64\times 64$ sub-scans from the patient's full scan that overlap at least in 10\% of their voxels with the corresponding artery segmentation mask. Such a restriction ensures that pre-training will happen on areas with higher likelihood of containing aneurysms. 
In addition to this choice, there are two key differences with respect to the original MAE setup. First, instead of masking patches with an uniform random distribution, we bias the selection toward patches close to blood arteries (see Fig. \ref{fig:pipe} (a)); motivation for this decision is that aneurysms are typically located in arteries, and by masking arteries out we force the model to learn better representations for such areas. 
Second, instead of only reconstructing scan intensity values, we concatenate artery distance maps to the scans as a second channel and train the encoder to reconstruct the distance values as well.
This forces the model to reason about the proximity of each reconstructed patch to the nearest arteries regardless of whether they are masked. Another benefit of reconstructing distance maps is that they can be used as an input channel during fine-tuning, making them useful across the pipeline.

\subsubsection{Finetuning strategy.}

For finetuning, we train both the encoder and detection module (without the decoder for MAE pre-training) on the annotated aneurysm detection data \cite{boHumanInterventionfreeClinical2021}, albeit using a lower learning rate compared with the pre-training stage. 
Our loss function is the average of 4 losses, similar to \cite{ceballos-arroyoVesselAwareAneurysmDetection2024}: binary cross entropy (BCE) for aneurysm classification and three independent MSE losses for localization, size, and intersection over union. For each ground truth aneurysm, we use Hungarian matching to pair it to the closest detection/query. The spatial losses are only computed when there is a positive match; otherwise, only the classification loss is used. To avoid situations where most of the queries are assigned to the background class (and thus only classification is optimized), we allow several detections to be matched to a given ground truth if they are close enough to it (under 1 mm). 

\section{Results}

\begin{table}[t!]
\fontsize{8pt}{9pt}\selectfont
\centering
\caption{ Detection performance with $t_{IoU}=0.3$ and assuming a fixed FPr=$0.5$.  We report {Se@FPr} curves to illustrate the threshold-agnostic performance of each model in Fig \ref{fig:froc}. Out-of-distribution (O.O.D.) datasets are highlighted. The private partition contains no healthy patients. }\label{tab:det1}
\begin{tabular}{|l|c|l|c|c|c|c|c|c|}
\hline
& &  & \multicolumn{4}{c|}{\textbf{Se~$\uparrow$} (\%)}  &  & \\
\textbf{Dataset} & \textbf{O.O.D.} & \textbf{Model} &\textbf{All}& \textbf{Small}&\textbf{ Med.} &\textbf{Large} &  \textbf{P-Se}~$\uparrow$ & \textbf{P-Sp}~$\uparrow$\\
\hline
\multirow{4}{*}{Bo Int. Test} & \multirow{4}{*}{\xmark} & CPM-Net & 87.3 & 72.4 &  93.3 & 71.4 &  86/102 &34/50\\
&&nnDet &  77.0 & 65.5 & 80.0 & 85.7 & 83/102  & 33/50 \\
&&Deform &  90.5 & \textbf{79.3} & 93.3 & \textbf{100} & 91/102  & 29/50 \\
 &&  Ours & \textbf{92.9} & 75.9 & \textbf{97.7} & \textbf{100}   & \textbf{93/102} &\textbf{ 35/50} \\
\hline 
\multirow{4}{*}{Bo Ext. Test}& \multirow{4}{*}{\cmark} & CPM-Net & 90.1 & 57.1& \textbf{95.9}& 92.8& 82/92 & \textbf{37/46} \\
&& nnDet & 88.1   & 64.3 & 90.4 & \textbf{100} & 84/92 & 30/46 \\
&&Deform &  89.1 & 64.3 & 93.1 & 92.8 & 82/92 & 35/46 \\
 &&  Ours & \textbf{93.1}& \textbf{71.4} & \textbf{95.9}& \textbf{100}   & \textbf{85/92} & 36/46 \\
\hline 
\multirow{4}{*}{CMHA}& \multirow{4}{*}{\cmark} & CPM-Net & 81.1  &63.9&\textbf{98.1}& 20.0 & 77/95 & \textbf{37/48} \\
&&nnDet & 84.2  & 66.6 & 94.4 & \textbf{100} &  80/95 & 40/48 \\
&&Deform &  82.1 & 58.3 & \textbf{98.1} & 80.0 &  78/95 & 31/48 \\
 &&  Ours &\textbf{90.5}& \textbf{80.5} & \textbf{98.1}& 80.0   & \textbf{86/95} & 35/48 \\
\hline 
\multirow{4}{*}{Private}& \multirow{4}{*}{\cmark} & CPM-Net & 57.5 & 39.28& 67.5& \textbf{80.0}& 75/143  &--  \\
&& nnDet & 59.4 & 44.0 & 68.3 & 73.3 & 73/143  & -- \\
&&Deform &  66.7 & 52.4 & 75.0 & \textbf{80.0 }&  86/143 & -- \\
 &&  Ours & \textbf{72.6} & \textbf{61.9} & \textbf{79.2}& \textbf{80.0 }  &  \textbf{94/143} & -- \\
\hline 
\end{tabular}
\end{table}

\subsubsection{Setup.}
We use compute nodes with 4 NVIDIA A100 80GB GPUs and 96 CPU cores each for training and a workstation with 32 CPU cores and a NVIDIA 4090 24GB GPU for processing private data. In our experiments, we use a Transformer encoder configuration with 6 layers, two sets of 8 attention heads each (one spatial, the other temporal), and a model dimension of 384, with a matching decoder for pre-training. Our tokenizer groups sets of $4\times 4\times 4$ voxels together, for a total of $16\times 16\times 16$ patches. We mask 75\% of the patches following \cite{heMaskedAutoencodersAre2022}. The detector uses 8 queries and a single transformer layer with 8 attention heads. During pre-training, we train for 100 epochs with the AdamW optimizer, using a learning rate of \num{1.5e-3} scheduled to decrease to \num{1.5e-4}. For fine-tuning we do 50 epochs with AdamW and a learning rate of \num{1e-4}. In both stages we use a weight decay of 0.05. We consider three strong baselines, two of which have achieved state-of-the-art results in aneurysm detection: nnDetection \cite{baumgartnerNnDetectionSelfconfiguringMethod2021a}, CPM-Net \cite{songCPMNet3DCenterPoints2020},  and the multi-scale deformable detector described in \cite{ceballos-arroyoVesselAwareAneurysmDetection2024}. nnDetection is a self-configuring pipeline which fits a Retina-UNet-based detector using 5-fold cross-validation for inference; CPM-Net is an anchor-free, CNN-based detector leveraging a 3D squeeze-and-excitation attention mechanism; 
and the deformable model consists of a CNN encoder which produces features for a multi-scale deformable attention mechanism leveraging vessel distance maps as a positional encoding. All three models were trained in a fully supervised manner on Bo \textit{et al}.'s dataset \cite{boHumanInterventionfreeClinical2021}.

\begin{figure}[t]
\centering
\includegraphics[width=1\textwidth]{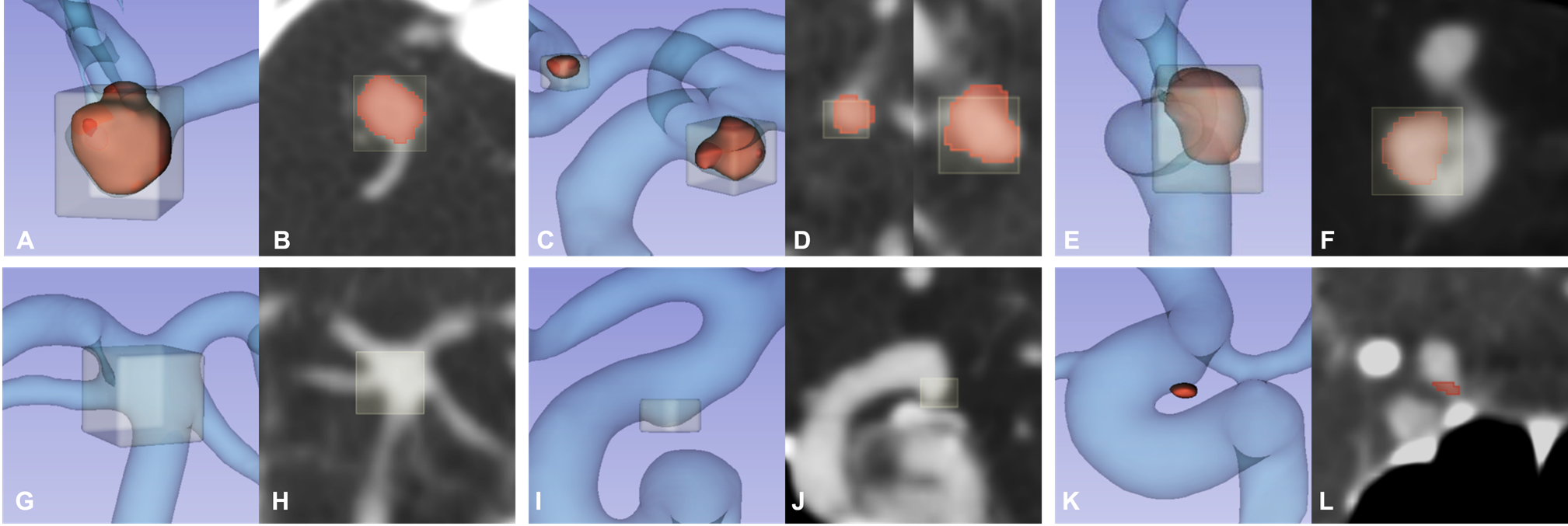}
\caption{ 3D view and corresponding CTA images. Red: ground-truth aneurysm; Yellow: algorithm output; Blue: artery segmentation. Top row (all TP): Right MCA aneurysm (A, B), anterior communicating artery aneurysm (smaller) and left posterior communicating artery aneurysm (larger) (C,D), left ICA aneurysm (E, F). Bottom row: FP-basilar tip confluence (G, H), FP-posterior communicating artery infundibulum (I, J), FN-small ICA aneurysm (K, L) }
\label{fig:qual} 
\end{figure}

\subsubsection{Evaluation metrics.}

Our main evaluation metric is Lesion-level Sensitivity (Se), \textit{i.e.}, the percentage of aneurysms correctly detected by a model. Based on conversations with radiologists, we determined an average of 0.5 FPs per scan to be the ideal tolerance at which to measure model reliability (as clinicians cannot be expected to filter through large amount of FPs), so Table \ref{tab:det1} only reports Sensitivity at such value. Nevertheless, Sensitivity vs FPr curves for all models on all datasets are shown in Fig. \ref{fig:froc}. Other reported metrics include Patient Level Sensitivity (P-Se) and Patient Level Specificity (P-Sp). To determine what constitutes a true positive (TP), we use an intersection over union (IoU) threshold of 0.3, which is at least as restrictive as those defined in the literature \cite{ceballos-arroyoVesselAwareAneurysmDetection2024,assisIntracranialAneurysmDetection2024}. In addition, we report Sensitivity results for small, medium, and large (diameters of 0-3 mm, 3-7 mm, and 7+ mm, respectively) aneurysms across datasets.

\begin{table}[h!]
\fontsize{8pt}{9pt}\selectfont
\centering
\caption{ Evaluation of the impact of our design choices on Specificity at a FP rate of 0.5 across 4 datasets. \textbf{A} is our best model; statistically significant results (via permutation testing) with respect to \textbf{A} are marked with an asterisk. }\label{tab:abl}
\begin{tabular}{|c|c|c|c|c|c|c|c|c|c|}
\hline
& \multicolumn{4}{c|}{\textbf{Pre-training strategy} }& \multicolumn{4}{c|}{\textbf{Se~$\uparrow$ (\%)} } \\
\textbf{Mod.} &\textbf{Mask.} & \textbf{Sampl.} & \textbf{Reco.} &\textbf{\# scans}& \textbf{Intern.} & \textbf{Extern. }& \textbf{CMHA} & \textbf{Private }\\
\hline
\textbf{A} &  \cmark & \cmark & \cmark & 6,796 & \textbf{92.9} & \textbf{93.1} & \textbf{90.5 }& \textbf{72.6}  \\
\textbf{B}& \cmark & \cmark  & \cmark & 3,500 & 88.9 & 90.1 & 81.1 * & 68.9 \\
\textbf{C} & \cmark &  \cmark & \cmark & 1,391 & 87.3 * & 84.2 * & 75.8 * & 63.5 * \\
\textbf{D }&  \cmark & \cmark & \xmark & 6,796 & 91.3 &\textbf{ 93.1} & 82.1 * & 70.8 \\
\textbf{E} &  \cmark & \xmark & \xmark & 6,796 & 82.5 * & 91.1 & 78.9 * & 58.0 * \\
\textbf{F} &  \xmark & \xmark & \xmark & 6,796 & 78.6 *& 79.2  *& 72.6 *& 53.4 *\\
\textbf{G} &  -- & -- & -- & -- & 80.2 * & 82.2 * & 76.8 * & 58.9 *\\
\hline 
\end{tabular}
\end{table}

\subsubsection{Comparisons with other methods.}
Table \ref{tab:det1} illustrates our results in contrast with the three baselines across four datasets and three aneurysm sizes. We observe that our approach consistently outperforms the others, with a wider gap when it comes to out of distribution (O.O.D.) data. The qualitative results depicted in Figure \ref{fig:qual} show that our model is able to accurately detect aneurysms of various sizes and located at various key locations in the brain vasculature. Our approach also produced several FPs, but we noted that these proved difficult for our radiologist colleagues to identify as such upon review. Some missed cases include very tiny aneurysms (Fig. \ref{fig:qual} K, L) which may be a result of the model using a single, somewhat coarse input resolution. This is mirrored in our model's results on the smallest aneurysms of Bo's internal test partition, where the deformable model outperforms it. Nevertheless, our implementation is efficient enough to explore more fine-grained resolutions in the future. Interestingly, all models perform significantly worse on our private dataset, which contains no healthy patients. This belies the need for training strategies that can adapt to populations whose distributions are drastically different from the training data. 

\subsubsection{Ablation studies.}
As for the impact of each design choice in our pipeline, Table \ref{tab:abl} depicts the effect of introducing more artery-based steps to MAE pre-training, as well as the impact of having more pre-training data. Across the board, the biggest impact results from sampling sub-scans from areas intersecting with vessels, although going from 1,391 to 6,796 pre-training scans results in significantly better results across the three O.O.D. datasets, which highlights the scalability of our approach. Notably, MAE pre-training without any of our changes results in worse performance when compared to fully supervised training.

\section{Conclusion}

In this paper, we described a novel approach for masked autoencoder pre-training toward aneurysm detection using artery information as a guidance mechanism for sampling, masking, and reconstruction. After fine-tuning, our model surpasses the previous SOTA models on the CT-based intracranial aneurysm detection task. Compared with such models, ours matches their performance on in-distribution data while achieving significantly better results on three out of distribution datasets, showcasing the impact of using unannotated data with anatomical guidance. Crucially, our results suggest that downstream performance scales with pre-training dataset size, highlighting there is room for further improvements. Moreover, unlike other approaches, we use a fully Transformer architecture with a factorized self-attention mechanism. In addition to being able to carry out MAE pre-training, using such an architecture provides ample opportunities for future work combining image and text modalities for more personalized clinical applications.

\begin{credits}
\subsubsection{\ackname} The authors acknowledge the financial support provided by NIH grant 1R01LM013891-01A1. 
Ceballos-Arroyo, A. and Jiang H. acknowledge the National Artificial Intelligence Research Resource (NAIRR) Pilot (NAIRR240236) and Microsoft Azure for contributing to this research result.
Ceballos-Arroyo, A. is grateful for the funding provided by Colombia's Minciencias and Fulbright under the Fulbright Minciencias 2021 program.

\subsubsection{\discintname}
The authors declare no competing interests.

\end{credits}

\bibliographystyle{splncs04}
\bibliography{refs}

\begin{thebibliography}{10}
\providecommand{\url}[1]{\texttt{#1}}
\providecommand{\urlprefix}{URL }
\providecommand{\doi}[1]{https://doi.org/#1}

\bibitem{arnabViViTVideoVision2021a}
Arnab, A., Dehghani, M., Heigold, G., Sun, C., Lu{\v c}i{\'c}, M., Schmid, C.: {{ViViT}}: {{A Video Vision Transformer}} (Nov 2021). \doi{10.48550/arXiv.2103.15691}

\bibitem{assisAneurysmPoseEstimation2023}
Assis, Y., Liao, L., Pierre, F., et~al.: Aneurysm {{Pose Estimation}} with {{Deep Learning}}. In: Greenspan, H., et~al. (eds.) Medical {{Image Computing}} and {{Computer Assisted Intervention}} -- {{MICCAI}} 2023, vol. 14221, pp. 543--553. Springer Nature Switzerland, Cham (2023). \doi{10.1007/978-3-031-43895-0_51}

\bibitem{assisIntracranialAneurysmDetection2024}
Assis, Y., Liao, L., Pierre, F., et~al.: Intracranial aneurysm detection: An object detection perspective. International Journal of Computer Assisted Radiology and Surgery  \textbf{19}(9),  1667--1675 (Sep 2024). \doi{10.1007/s11548-024-03132-z}

\bibitem{baumgartnerNnDetectionSelfconfiguringMethod2021a}
Baumgartner, M., J{\"a}ger, P.F., Isensee, F., {Maier-Hein}, K.H.: {{nnDetection}}: {{A Self-configuring Method}} for {{Medical Object Detection}}. In: Medical {{Image Computing}} and {{Computer Assisted Intervention}} -- {{MICCAI}} 2021. pp. 530--539. Lecture {{Notes}} in {{Computer Science}}, {Springer International Publishing}, {Cham} (2021)

\bibitem{boHumanInterventionfreeClinical2021}
Bo, Z.H., Qiao, H., Tian, C., Guo, Y., et~al.: Toward human intervention-free clinical diagnosis of intracranial aneurysm via deep neural network. Patterns  \textbf{2}(2),  100197 (Feb 2021). \doi{10.1016/j.patter.2020.100197}

\bibitem{detr}
Carion, N., Massa, F., Synnaeve, G., Usunier, N., Kirillov, A., Zagoruyko, S.: End-to-end object detection with transformers. In: European Conference on Computer Vision (ECCV) (2020)

\bibitem{caronEmergingPropertiesSelfSupervised2021b}
Caron, M., Touvron, H., Misra, I., et~al.: Emerging {{Properties}} in {{Self-Supervised Vision Transformers}} (May 2021). \doi{10.48550/arXiv.2104.14294}

\bibitem{ceballos-arroyoVesselAwareAneurysmDetection2024}
{Ceballos-Arroyo}, A.M., Nguyen, H.T., Zhu, F., et~al.: Vessel-{{Aware Aneurysm Detection Using Multi-scale Deformable 3D Attention}}. In: Linguraru, M.G., et~al. (eds.) Medical {{Image Computing}} and {{Computer Assisted Intervention}} -- {{MICCAI}} 2024. pp. 754--765. Springer Nature Switzerland, Cham (2024). \doi{10.1007/978-3-031-72086-4_71}

\bibitem{chenDevelopmentQuantitativeIntracranial2018}
Chen, L., {Mossa-Basha}, M., Balu, N., et~al.: Development of a {{Quantitative Intracranial Vascular Features Extraction Tool}} on {{3D MRA Using Semi-automated Open-Curve Active Contour Vessel Tracing}}. Magnetic resonance in medicine  \textbf{79}(6),  3229--3238 (Jun 2018). \doi{10.1002/mrm.26961}

\bibitem{chilamkurthyDeepLearningAlgorithms2018}
Chilamkurthy, S., Ghosh, R., Tanamala, S., et~al.: Deep learning algorithms for detection of critical findings in head {{CT}} scans: A retrospective study. The Lancet  \textbf{392}(10162),  2388--2396 (Dec 2018). \doi{10.1016/S0140-6736(18)31645-3}

\bibitem{congSatMAEPretrainingTransformers2023}
Cong, Y., Khanna, S., Meng, C., Liu, P., Rozi, E., He, Y., Burke, M., Lobell, D.B., Ermon, S.: {{SatMAE}}: {{Pre-training Transformers}} for {{Temporal}} and {{Multi-Spectral Satellite Imagery}} (Jan 2023). \doi{10.48550/arXiv.2207.08051}

\bibitem{dengGeneralizationAbilityUnsupervised2024}
Deng, Y., Hong, J., Zhou, J., Mahdavi, M.: On the {{Generalization Ability}} of {{Unsupervised Pretraining}}. Proceedings of machine learning research  \textbf{238},  4519--4527 (May 2024)

\bibitem{dinotoAutomatedBrainAneurysm2023}
Di~Noto, T., Marie, G., Tourbier, S., et~al.: Towards {{Automated Brain Aneurysm Detection}} in {{TOF-MRA}}: {{Open Data}}, {{Weak Labels}}, and {{Anatomical Knowledge}}. Neuroinformatics  \textbf{21}(1),  21--34 (Jan 2023). \doi{10.1007/s12021-022-09597-0}

\bibitem{dosovitskiyImageWorth16x162021a}
Dosovitskiy, A., Beyer, L., Kolesnikov, A., et~al.: An {{Image}} is {{Worth}} 16x16 {{Words}}: {{Transformers}} for {{Image Recognition}} at {{Scale}} (Jun 2021). \doi{10.48550/arXiv.2010.11929}

\bibitem{feichtenhoferMaskedAutoencodersSpatiotemporal2022}
Feichtenhofer, C., Fan, H., Li, Y., He, K.: Masked {{Autoencoders As Spatiotemporal Learners}} (Oct 2022). \doi{10.48550/arXiv.2205.09113}

\bibitem{heMaskedAutoencodersAre2022}
He, K., Chen, X., Xie, S., Li, Y., Dollar, P., Girshick, R.: Masked {{Autoencoders Are Scalable Vision Learners}}. In: 2022 {{IEEE}}/{{CVF Conference}} on {{Computer Vision}} and {{Pattern Recognition}} ({{CVPR}}). pp. 15979--15988. IEEE, New Orleans, LA, USA (Jun 2022). \doi{10.1109/CVPR52688.2022.01553}

\bibitem{hoellCorticalVenousAneurysm2004}
Hoell, T., Hohaus, C., Beier, A., Holzhausen, H.J., Meisel, H.J.: Cortical {{Venous Aneurysm Isolated Cerebral Varix}}. Interventional Neuroradiology  \textbf{10}(2), ~161 (Oct 2004). \doi{10.1177/159101990401000210}

\bibitem{hooperImpactUpstreamMedical2021a}
Hooper, S.M., Dunnmon, J.A., Lungren, M.P., Mastrodicasa, D., et~al.: Impact of {{Upstream Medical Image Processing}} on {{Downstream Performance}} of a {{Head CT Triage Neural Network}}. Radiology: Artificial Intelligence  \textbf{3}(4),  e200229 (Jul 2021)

\bibitem{isenseeNnUNetSelfconfiguringMethod2021}
Isensee, F., Jaeger, P.F., Kohl, S.A.A., et~al.: {{nnU-Net}}: A self-configuring method for deep learning-based biomedical image segmentation. Nature Methods  \textbf{18}(2),  203--211 (Feb 2021). \doi{10.1038/s41592-020-01008-z}

\bibitem{khoruzhayaExpandedBrainCT2024}
Khoruzhaya, A.N., Bobrovskaya, T.M., Kozlov, D.V., et~al.: Expanded {{Brain CT Dataset}} for the {{Development}} of {{AI Systems}} for {{Intracranial Hemorrhage Detection}} and {{Classification}}. Data  \textbf{9}(2), ~30 (Feb 2024)

\bibitem{Park_Lee_Heo_Han_Lee_Hong_Lee_Lee_Oh_2022}
Park, S.W., Lee, J.Y., Heo, N.H., et~al.: Short- and long-term mortality of subarachnoid hemorrhage according to hospital volume and severity using a nationwide multicenter registry study. Frontiers in Neurology  \textbf{13},  952794 (Aug 2022)

\bibitem{ridnikImageNet21KPretrainingMasses2021}
Ridnik, T., {Ben-Baruch}, E., Noy, A., {Zelnik-Manor}, L.: {{ImageNet-21K Pretraining}} for the {{Masses}} (Aug 2021). \doi{10.48550/arXiv.2104.10972}

\bibitem{shiClinicallyApplicableDeeplearning2020}
Shi, Z., Miao, C., Schoepf, U.J., et~al.: A clinically applicable deep-learning model for detecting intracranial aneurysm in computed tomography angiography images. Nature Communications  \textbf{11}(1), ~6090 (Nov 2020)

\bibitem{songIntracranialAneurysmCTA2024}
Song, M., Wang, S., Qian, Q., Zhou, Y., Luo, Y., Gong, X.: Intracranial aneurysm {{CTA}} images and {{3D}} models dataset with clinical morphological and hemodynamic data. Scientific Data  \textbf{11}(1), ~1213 (Nov 2024). \doi{10.1038/s41597-024-04056-8}

\bibitem{songCPMNet3DCenterPoints2020}
Song, T., Chen, J., Luo, X., et~al.: {{CPM-Net}}: {{A 3D Center-Points Matching Network}} for {{Pulmonary Nodule Detection}} in {{CT Scans}}. In: Martel, A.L., et~al. (eds.) Medical {{Image Computing}} and {{Computer Assisted Intervention}} -- {{MICCAI}} 2020, vol. 12266, pp. 550--559. Springer International Publishing, Cham (2020)

\bibitem{tongVideoMAEMaskedAutoencoders2022}
Tong, Z., Song, Y., Wang, J., Wang, L.: {{VideoMAE}}: {{Masked Autoencoders}} are {{Data-Efficient Learners}} for {{Self-Supervised Video Pre-Training}} (Oct 2022). \doi{10.48550/arXiv.2203.12602}

\bibitem{wangDetectionIntracranialAneurysms2023}
Wang, J., Sun, J., Xu, J., et~al.: Detection of {{Intracranial Aneurysms Using Multiphase CT Angiography}} with a {{Deep Learning Model}}. Academic Radiology  \textbf{30}(11),  2477--2486 (Nov 2023). \doi{10.1016/j.acra.2022.12.043}

\bibitem{yadavDynamicComputedTomographyAngiography2025a}
Yadav, S., Kim, J., Young, G., Qin, L.: Dynamic-{{Computed Tomography Angiography}} for {{Cerebral Vessel Templates}} and {{Segmentation}} (Feb 2025). \doi{10.48550/arXiv.2502.09893}

\bibitem{yangBenchmarkingCoWTopCoW2024}
Yang, K., Musio, F., Ma, Y., et~al.: Benchmarking the {{CoW}} with the {{TopCoW Challenge}}: {{Topology-Aware Anatomical Segmentation}} of the {{Circle}} of {{Willis}} for {{CTA}} and {{MRA}} (Apr 2024). \doi{10.48550/arXiv.2312.17670}

\bibitem{zhouSelfPretrainingMasked2023}
Zhou, L., Liu, H., Bae, J., He, J., Samaras, D., Prasanna, P.: Self {{Pre-training}} with {{Masked Autoencoders}} for {{Medical Image Classification}} and {{Segmentation}} (Apr 2023). \doi{10.48550/arXiv.2203.05573}

\bibitem{zhuDeformableDETRDeformable2021}
Zhu, X., Su, W., Lu, L., et~al.: {{Deformable Transformers}} for {{End-to-End Object Detection}} (2021). \doi{10.48550/arXiv.2010.04159}

\end{thebibliography}

\end{document}